# High-Order Question Generation in a Multilingual Educational Context

**Suna-Şeyma Uçar[1], Itziar Aldabe[1], Nora Aranberri[1], Orphée De Clercq[2]**
[1]HiTZ Basque Center for Language Technology
Ixa NLP Group, University of the Basque Country UPV/EHU
[2]LT[3], Language and Translation Technology Team, Ghent University, Ghent, Belgium
sunaseyma.ucar@ehu.eus, itziar.aldabe@ehu.eus,
nora.aranberri@ehu.eus, orphee.declercq@ugent.be

**Abstract**

Critical thinking is a fundamental skill that helps learners move beyond simple memorization. One way to develop this skill is through high-order questioning. However, crafting such questions remains a challenge for educators, and classroom practices tend to rely on low-order questions. Large Language Models have demonstrated strong capabilities in generating high-order questions, especially when guided by prompts based on Bloom's Taxonomy. Yet, existing research has largely centered on this framework and focused only on English. This study addresses these gaps by introducing prompts grounded in two alternative frameworks: Claim-Evidence-Reasoning and Divergent Questioning within a multilingual context using Basque, Spanish, and English. Results indicate that while both an open-source and a proprietary model rather effectively generate questions in all three languages, only about half of the answerable questions are recognized by teachers as high-order. A positive finding is that the alternative frameworks produce structurally and conceptually varied questions, suggesting they could complement each other and provide viable alternatives to Bloom's Taxonomy.

## 1. Introduction

In the learning process, critical thinking shapes a learner's ability to analyze, evaluate, and apply knowledge. Kurfiss (1988, p. 2) defines it as "an investigation whose purpose is to explore a situation, phenomenon, question, or problem to arrive at a hypothesis or conclusion that integrates all available information and can be convincingly justified."

Research emphasizes that questioning has the greatest impact on students' thinking, and that strategies are essential to challenge and extend it (Clasen and clasen, 1990; Diaz et al., 2013). A key skill for fostering critical thinking is a teacher's ability to ask high-order questions, defined by Bloom's Revised Taxonomy (Krathwohl, 2002) (hereafter referred to as Bloom's Taxonomy), a widely used educational framework for classifying learning objectives according to cognitive complexity, which categorizes cognitive processes into six hierarchical levels. The top three levels (analyze, evaluate, create) correspond to high-order levels which require students to acquire analysis, evaluation and creation skills.

High-order questions require complex reasoning and critical assessment. Savage (1998) demonstrated that students forget 80–90% of information from factual questions but retain 80–85% from high-order questions. Despite their importance, classroom questions often remain at lower cognitive levels (Pate and Bremer, 1967; Gall, 1970). Practical constraints and the expertise required to craft the questions hinder their use. Automated question generation (QG) via generative AI has emerged as a promising solution (Elkins et al., 2024).

Our research investigates whether Large Language Models (LLMs) can support educators by producing high-order questions aligned with three educational frameworks: Bloom's Taxonomy, Claim-Evidence-Reasoning (CER) (McNeill et al., 2006) and divergent questioning (Gallagher and Aschner, 1963). While Bloom's Taxonomy has been studied (Elkins et al., 2023), CER and divergent questioning offer complementary, structured approaches to fostering critical thinking, through evidence-based reasoning and the consideration of multiple perspectives, which we believe can also support the generation of higher-order questions. Unlike prior work focused on English, we address multilingual contexts, generating high-order physics questions in Basque, Spanish, and English from secondary school texts. Experiments were conducted with two state-of-the-art LLMs: the open-source Llama3_70b (Llama3) and the proprietary GPT-4. Generated questions were manually analyzed in two ways: teachers evaluated them for cognitive demand and pedagogical usefulness, and answerable high-order questions were further examined for specific content and form-related characteristics.

Our results show that most generated questions were answerable, though there is still room for improvement in model performance. Over half of these were classified by teachers as high-order, highlighting the potential of the tested frameworks to guide models in this task and the benefit of exploring alternative approaches. Prompts based on Bloom's Taxonomy Evaluate and divergent questioning were more effective to generate high-order questions. Both the CER and divergent questioning strategies produced questions with characteristics complementary to Bloom's Taxonomy. Both LLMs generated questions in all three languages, with GPT-4 performing slightly better in Spanish and English, and Llama3 in Basque.

The remainder of this paper is structured as follows: Section 2 provides an overview of the literature on the evolution of QG, particularly in the context of modern LLMs. Section 3 outlines the methodology and experiments conducted. Section 4 presents the evaluation results from secondary school teachers, including response distributions and inter-annotator agreement outcomes. Finally, Section 6 discusses the results and offers practical implications for using LLMs for QG.

## 2. Related Work

Recently, LLMs have captured attention in educational research, particularly for QG using various prompting strategies. Given our focus on generating high-order questions, we center this section on studies examining QG with LLMs and educational frameworks.

Previous studies explored the QG task using frameworks such as Bloom's Taxonomy. Elkins et al. (2023) examined LLMs' effectiveness in generating English classroom questions with few-shot prompting, using InstructGPT with both Bloom's and a difficulty taxonomy (easy, medium, hard). Experiments in the ML and biology domains at university level used 68 Wikipedia passages, generating 612 questions across Bloom's and difficulty levels. Teachers evaluated them on relevance, adherence, grammar, answerability, and usefulness. Results showed high quality in relevance, grammar, and answerability, concluding that InstructGPT effectively supports teachers, especially for low-order questions.

Elkins et al. (2024) utilized GPT-3.5 for generating English quiz questions aligned with Bloom's Taxonomy, comparing simple (generic) and controlled (structured) prompting. Using 24 Wikipedia passages from ML and biology, teachers produced handwritten, simple, and controlled quizzes. Simple prompts generated six questions per context, while controlled prompts applied Bloom's Taxonomy with five-shot learning. Eight evaluators assessed quizzes on nine metrics, including coverage, structure, and answerability. Results showed LLM-generated quizzes matched handwritten quality, with controlled prompting preferred, indicating LLMs can aid teachers in producing questions.

Scaria et al. (2024) also explored QG based on Bloom's Taxonomy, utilizing Mistral, Llama2, PaLM 2, GPT-3.5, and GPT-4 to generate English questions across cognitive levels. Five prompting strategies were used: a basic prompt; chain of thought (CoT) with Bloom's definitions; CoT with example questions; CoT with skill explanations and examples; and CoT combining all three. A total of 510 questions were generated and evaluated both manually and automatically. Two data science experts conducted human evaluation using a nine-item rubric for understandability, topic relatedness, and grammaticality. Evaluations focused on quality and adherence to Bloom's Taxonomy, concluding that strategic prompting enables LLMs to generate high-quality questions across Bloom's Taxonomy levels.

While prior research shows LLMs' potential for educational QG, it mainly focuses on English and higher education. We expand this by exploring multiple languages of instruction, namely, English, Spanish, and Basque, and by incorporating CER and divergent questioning frameworks as guides to generate high-order questions going beyond Bloom's Taxonomy.

## 3. Methodology

This study explores LLMs' capacity to generate high-order questions in order to foster critical thinking skills in classroom settings. Building on prior research using Bloom's Taxonomy as a foundation for prompting strategies (Scaria et al., 2024), we adapted it to a document-level context, focusing on the high-order levels. Bloom's Taxonomy defines six cognitive levels: remembering, understanding, applying, analyzing, evaluating, and creating. The first three can be considered as low-order and the latter three as high-order.

We employ the CER framework, which emphasizes reasoning by guiding students to identify a central claim, provide supporting evidence, and logically connect the evidence to the claim (McNeill et al., 2006). This framework is particularly well-suited for engaging with complex scientific concepts, making it a strong candidate for high-order questioning.

In addition to Bloom's Taxonomy and the CER framework, a third framework focuses on divergent questioning, which promotes exploration and creativity by encouraging questions with multiple potential answers or interpretations (McNamara, 1981). Divergent questions stimulate discussion and critical thinking, aligning well with the goals of high-order cognitive development.

We used a consistent prompt structure across all three frameworks, adapting the top-performing prompt design from Scaria et al. (2024). This de-sign incorporates a CoT strategy, providing models with detailed guidance. Each prompt includes: (1) a brief description of the intended audience, (2) a summary of the educational framework, (3) a request for questions tailored to Spanish secondary school students, and (4) the STEM text for QG. An example of the prompt template is provided below:

---

**Prompt Template:**
*You are an experienced secondary school teacher in STEM subjects. [Educational framework] is a structured approach to [purpose of framework]. [Brief description of the framework's key elements or goals]. Based on this framework, write a question that asks students to [cognitive skill or purpose of the framework] related to the content provided in the text.*
Please make sure that the questions are relatable to students in Spain.
This is the text: {text}

---

The study is situated within the Basque educational system, where English, Spanish, and Basque are languages of instruction. We tested our approach across all three languages using 30 secondary school-level physics texts (10 per language) for students aged 12–16. Physics was chosen to minimize bias from domain-specific knowledge. Texts were sourced from the Agrega2 project[1], a national initiative co-funded by the EU (Feder) providing an online repository of educational resources. This corpus has been used in multilingual readability studies (Uçar et al., 2024) and is tailored to students in the Basque Autonomous Community (BAC).

For QG, we utilized two advanced LLMs available at the time of the study: Llama3.1_70B (open-source) and GPT-4 (proprietary). Both models have demonstrated strong performance across a variety of educational tasks, including QG (Elkins et al., 2023). Each model was prompted to generate five questions for each of the 30 texts, resulting in 300 questions in total.

### 3.1. Evaluation

Our research question explores whether apart from Bloom's Taxonomy, two additional frameworks (CER and Divergent Questioning) can be used as part of a prompting strategy to generate high-order questions. To address this question, two manual evaluations were conducted.

For the first evaluation, secondary school teachers from an educational institution in the BAC were recruited. The participating teachers specialized in natural science subjects and were proficient in Basque (L1), Spanish (L1), and English (L2). A total of four teachers participated: two with over 10 years of teaching experience and two with 3 and 2 years of experience, respectively. The 300 questions were evenly distributed among three evaluators. To assess inter-annotator agreement (IAA), the fourth evaluator independently reviewed a subset of 90 questions, i.e., 30 per language.

Teachers were provided with detailed guidelines outlining the evaluation task, including definitions of the Bloom's Taxonomy (BT) levels and descriptions of the additional evaluation metrics. Four evaluation metrics were employed: The **Answerability** metric determines whether the question can reasonably be answered by secondary school students based on the provided text or their current knowledge (*Yes*|*No*). If *No* is selected, participants are instructed to ignore the other metrics and move on to the evaluation of the next question. **Relevance** measures how well the generated questions align with the key ideas and content that students are expected to focus on in the text (*Not at all*|*Slightly*|*Moderately*|*Completely*). The **Bloom's Taxonomy Level** metric evaluates which level of BT the question targets (*Remember*|*Understand*|*Apply*|*Analyze*|*Evaluate*|*Create*). Finally, the **Teacher Adoption** metric evaluates the likelihood that these questions would be integrated into classroom instruction by the teachers (*No*|*Yes, with modifications*|*Yes, as is*).

For the second evaluation, all questions labeled as high-order by the teachers were further analyzed by the authors using six predefined criteria: **1 - Task type:** the kind of task the student is being asked to perform. **2 - Single and nested questions:** whether a question requires a single answer or multiple sub-answers. **3 - Lexical and structural patterns:** grammatical structures and phrasings. **4 - Prompt wording alignment:** whether the question incorporates specific key terms from the educational framework in the prompt. **5 - Text referencing:** whether the question explicitly mentions the source text. **6 - Question length:** whether different prompting strategies or LLMs influenced question length.

## 4. Results of Teacher Evaluation

### 4.1. Inter-annotator Agreement

For each metric, we report agreement rates along with Cohen's Kappa[2] and GWET AC1/AC2. The results (Table 1) reveal that this is a challenging task. Clearly these metrics can be considered highly subjective and be influenced by each teacher's background and experience. Moreover, teaching styles and classroom needs vary among teachers.

[1] http://www.agrega2.es/web/

[2] Please note that for the ordinal metrics quadratic weighted kappa was calculated.

| Metric | Agreement Rate | Cohen's Kappa | Gwet's AC1/AC2 |
|---|---|---|---|
| Answerability | .744 | .109 | .643 |
| Relevance | .556 | .291 | .264 |
| Bloom's Tax. | .365 | .380 | .266 |
| Adoption | .571 | .231 | .409 |

Table 1: Inter-annotator agreement (IAA) results. Agreement Rate is the proportion of identical annotations. Cohen's Kappa and Gwet's AC1/AC2 are chance-corrected agreement coefficients, where higher values indicate stronger agreement.

However, we also seem to observe a conflicting pattern between the raw agreement rates and Kappa scores. Kappa measures rater agreement while accounting for chance, but since it depends on category distribution, a high prevalence of any category can lower the Kappa value despite high agreement rates (Feinstein and Cicchetti, 1990). To address this limitation, alternative measures such as Gwet's AC1/AC2 have been proposed as more reliable chance-corrected agreement coefficients (Gwet, 2008). We assess the Kappa paradox using Gwet's AC1/AC2 values, calculating AC1 for Answerability and Adoption and AC2 for Relevance and Bloom's Taxonomy (ordinal values). As shown in Table 1, the Kappa paradox seems to be present in both Answerability and Adoption, where Cohen's Kappa remains low despite high agreement rates.

## 4.2. Response Distributions

The results presented in Table 2 indicate that nearly 77% of the generated questions were considered **Answerable** based on the accompanying text. For this subset of answerable questions, further evaluation shows that questions were distributed across all levels of **Bloom's Taxonomy**, with slightly more than half (50.41%) of the questions classified as high-order questions (i.e., Analyze, Evaluate or Create).

When it comes to a question's **Relevance**, almost 95.72% of the questions are considered completely relevant. Finally, in the **Teacher Adoption** metric, the responses are highly skewed towards the *Yes* categories (*Yes, with modifications* and *Yes, as is*), amounting to no less than 94.43%.

**Distributions across educational frameworks:**

Table 3 summarizes which Bloom's Taxonomy classification was indicated by the teachers for each question generated with each one of the three educational frameworks. We observe that the divergent questioning (69.81%) and Bloom's Taxonomy Evaluate (64.71%) prompts were most successful in generating high-order questions. Regarding

| Metric | Response | % |
|---|---|---|
| Answerability | No | 23.27 |
| | Yes | 76.72 |
| Relevance | Slightly | 4.27 |
| | Moderately | 20.51 |
| | Completely | 75.21 |
| Bloom's Taxonomy | Remember | 2.56 |
| | Understand | 31.62 |
| | Apply | 15.38 |
| | Analyze | 27.77 |
| | Evaluate | 14.95 |
| | Create | 7.69 |
| Teacher Adoption | No, I would not use it | 5.55 |
| | Yes, with modifications | 36.32 |
| | Yes, as is | 58.11 |

Table 2: Distribution of the LLM-generated question evaluations across four metrics: Answerability, Relevance, Bloom's Taxonomy level, and Teacher Adoption. Percentages represent the proportion of questions assigned to each response label.

Bloom's Taxonomy Analyze only 38.18% of the questions were considered high-order. A similar percentage is observed when relying on the CER framework (41.82%). With 52.94% the Bloom's Taxonomy Create strategy, on the other hand, also seems to result in more high-order questions, making it the third best strategy. What also draws attention is that only two prompts succeeded in creating high-order questions of the highest level (i.e., the *Create* level). Unsurprisingly, this level of questions was achieved most frequently using Bloom's Taxonomy Create strategy (35.29%). Also here, we observe that the divergent questioning prompt strategy is quite successful with 11.32% Create questions. Looking at the low-order questions we see that the Bloom's Taxonomy Analyze and CER groundings in the prompt substantially generate questions in the *Understand* level.

| | | Bloom's Tax. | | | CER | Div |
|---|---|---|---|---|---|---|
| | | Ana. | Eva. | Cre. | | |
| High | Create | 0 | 0 | **35.29** | 0 | 11.32 |
| | Evaluate | 1.82 | 39.22 | 5.88 | 7.27 | **41.51** |
| | Analyze | **36.36** | 25.49 | 11.76 | **34.55** | 16.98 |
| | Total | 38.18 | 64.71 | 52.94 | 41.82 | 69.81 |
| Low | Apply | 12.73 | 9.80 | 32.35 | 7.27 | 16.98 |
| | Understand | **43.64** | 23.53 | 14.71 | **47.27** | 13.21 |
| | Remember | 5.45 | 1.96 | 0 | 3.64 | 0 |
| | Total | 61.82 | 35.29 | 47.06 | 58.18 | 30.19 |

Table 3: Bloom's Taxonomy level distribution (%) of generated questions per prompting strategy (Bloom's Taxonomy [Analyze, Evaluate, Create], CER, and Divergent Questioning). Bold values indicate the highest percentage within each column; totals reflect the % of high-order and low-order questions.

**Distributions across LLMs:**

In Figure 1 we present the response distribution across each LLM for each evaluation metric and observe that Llama3 and GPT-4 exhibit similar trends. GPT-4 shows a slightly higher **Answerability** rate (79.3%) than Llama3 (74.7%). Both models display comparable **Relevance** trends, with most responses marked as completely relevant, with a slight preference for Llama3 (76.8%) compared to GPT-4 (73.1%). The same goes for the **Teacher Adoption** rates, where Llama33 is slightly more successful at generating questions which can be adopted as such (59.8%) compared to GPT-4 (57.1%). Looking at **Bloom's Taxonomy** levels we observe that all levels are present in the questions generated by both LLMs, but GPT-4 seems slightly more capable of generating high-order questions of the levels Analyze and Evaluate.

**Distributions across languages:**

Given that questions were generated in three different languages, it is interesting to also consider the results taking into account both the language and model-wise distributions, as presented in Figure 2. In English, GPT-4 outperforms Llama3 in **Answerability**, achieving 88.0%, compared to 80.0% for Llama3. For **Relevance**, both models obtain very similar results for *Completely Relevant* responses (72.7% for GPT-4 and 70% for Llama3). **Bloom's Taxonomy** results show that GPT-4's responses are more spread across different categories, with the **Analyze** level (40.9%) being most common, followed by **Evaluate** (22.7%). Llama3 obtains 30% for the **Understand** level followed by **Analyze** (22.5%). In English, GPT-4 clearly outperforms all other settings when it comes to high-order QG.

In Spanish, GPT-4 performs somewhat better than Llama3 in **Answerability** with models achieving 86% and 80.0%, respectively. In **Relevance**, Llama3 scores slightly better at 72.5% while GPT-4 obtained 67.4% at *Completely Relevant*. Llama3 has a lower percentage of less relevant responses (2.5% at *Slightly Relevant* vs. GPT-4's 9.3%). For **Teacher Adoption**, both models perform virtually the same at *Yes, as is* with 51.2% and 50.0%. However, Llama3 has a higher percentage at *Yes, with modifications* (47.5%) compared to 41.9% for GPT-4. Regarding **Bloom's Taxonomy** GPT-4 (46.5%) is slightly more successful in generating high-order questions in Spanish than Llama3 (40%).

Finally, for Basque we observe that regarding **Answerability** both models struggle more compared to English and Spanish, both achieving 64.0%. **Relevance** scores in Basque favor Llama3, which achieves 90.6% on the *Completely Relevant* metric, compared to GPT-4 with 81.2%. For **Teacher Adoption**, Llama3 clearly outperforms GPT-4, with 71.9% of responses rated as *Yes, as it is* compared to 59.4% for GPT-4. GPT-4 has a higher percentage of responses rated as *Yes, with modifications* (34.4%) than Llama3 (18.8%). Considering **Bloom's Taxonomy** levels, GPT-4 (43.70%) provides slightly more high-order questions than Llama3 (37.4%).

## 5. Results of High-Order Question Characteristics

Out of the 231 answerable questions, a total of 117 questions (26 in Basque, 36 in Spanish, and 55 in English) were classified as high-order by the teachers during the first evaluation step. We now present the results of the second manual analysis that further examines the content and form of the questions that were deemed adequate. We report the observation per language, each time zooming in on the three different frameworks and listing examples accompanying relevant criteria.

### 5.1. Basque

The questions generated with **Bloom's Taxonomy prompts**, at the Evaluate level, required students to make judgments or assess content, while Create-level questions asked students to design or imagine new ideas, often starting with verbs such as *diseinatu* (design) or *imajinatu* (imagine). Questions were mostly nested, asking students to perform more than one task. The questions tended to include the specific words in the prompts while no explicit reference was made to the texts. Across all levels, the majority of questions presented language-related issues such as incorrect placement of question words and mismatched grammatical forms Q1. Question length varied across prompts and LLMs: Llama3 generally produced shorter questions, while GPT-4 questions were longer and more varied. At the Create level, both models produced longer questions, the difference in length between them being smaller.

In the questions generated using the **CER**-based prompt, all questions were single questions, with none requiring multiple responses. Six out of seven questions included the key terms claim, evidence, and reasoning, as can be seen in Q2, indicating little deviation from the definition of the framework. Again, all the questions showed language problems. The questions did not make explicit reference to the source text. Again, in terms of length, GPT-4 generated longer questions than LLaMA3.

Although only two answerable high-order questions were generated using the prompt grounded in the **divergent questioning** framework, both were considered engaging and capable of eliciting diverse responses from students. Unfortunately, both exhibited language issues. Their lengths varied, with one being notably longer.

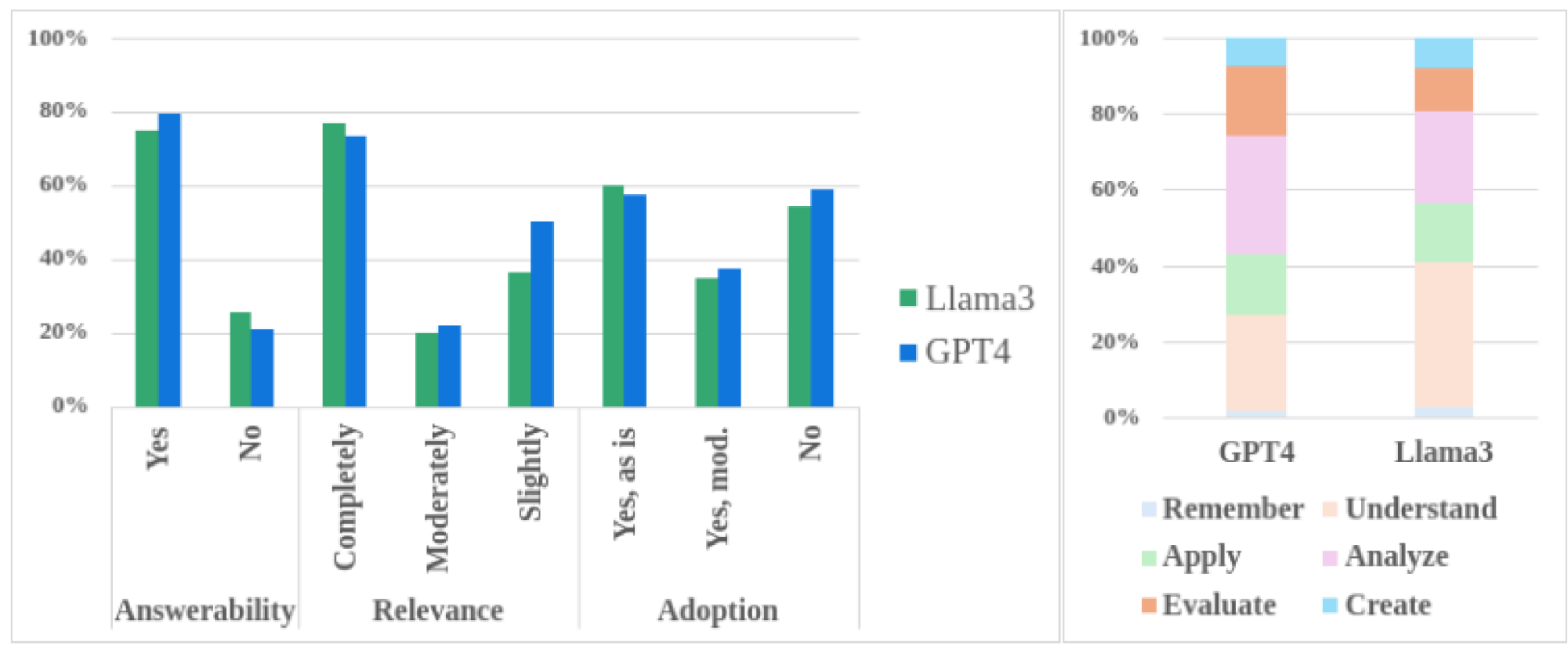

Figure 1: Evaluation metric distributions for GPT-4 and Llama3 generated questions. Left: response distributions for Answerability, Relevance, and Teacher Adoption. Right: Bloom's Taxonomy level distributions.

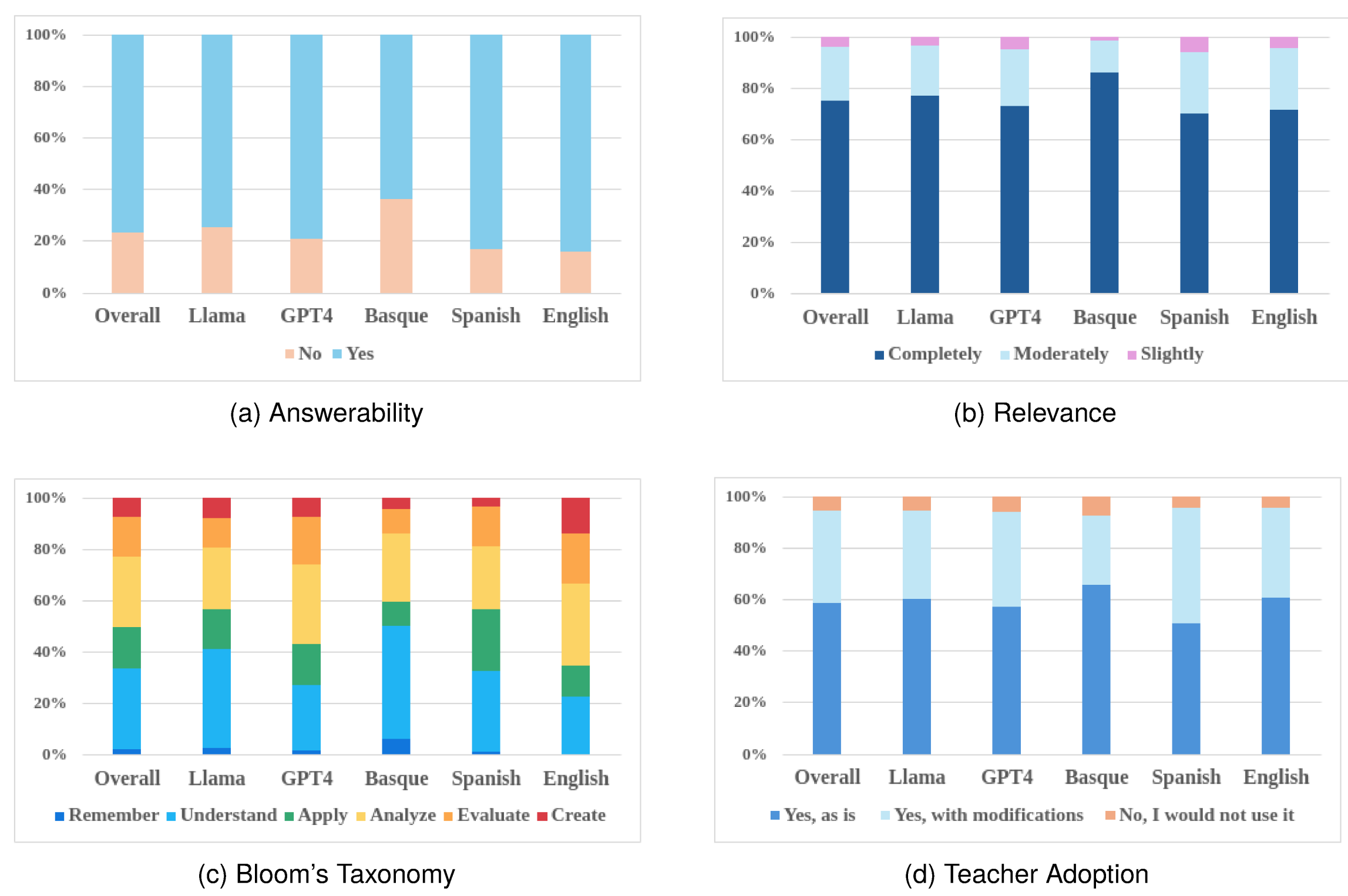

(a) Answerability

(b) Relevance

(c) Bloom's Taxonomy

(d) Teacher Adoption

Figure 2: Per-language and per-model response distributions for GPT-4 and Llama3 across four evaluation metrics: Answerability (top-left), Relevance (top-right), Bloom's Taxonomy levels (bottom-left), and Teacher Adoption (bottom-right). Results are broken down across three languages: English, Spanish, and Basque.

## 5.2. Spanish

Spanish questions based on **Bloom's Taxonomy** showed diversity in tasks depending on the level. At Analyze, 4 out of 5 questions began with emph¿Cómo? (How?) and 3 asked students to identify relationships between concepts (Q3). Evaluate questions showed more diversity; 4 out of 11 asked about the importance of a concept, and others asked for justification or comparison. For Create, 5 out of 6 questions asked students to design something, usually beginning with *Diseña* (Design) (Q4). Nested questions were common: 2 Analyze, 7 Evaluate, and 3 Create questions required students to respond to more than one question. Lexical and structural patterns varied across lev-els. All Analyze questions shared similar phrasings while Evaluate and Create questions featured more variation. Prompt wording was present in the questions. The source text was hardly ever referenced. Question lengths varied: Analyze questions were shortest, while Create questions were longer.

| | Question |
|---|---|
| Q1 | *Unibertsoaren osagaiak zer dira eta nola erlazionatzen dira? (What are the components of the universe and how do they relate? (Llama3))* |
| Q2 | *Baieztatu zein den unibertsoaren tamaina neurtzeko erabilitako neurtzen unitatea eta azaldu zergatik erabiltzen den unitate hori. Testuko frogak eta arrazoiketak erabil ditzakezu zure baieztapena defendatzeko. (Confirm which is the unit of measurement used to measure the size of the universe and explain why this unit is used. You can use the evidence and reasoning in the text to defend your claim.) (GPT-4)* |

Table 4: Questions generated with Llama3 and GPT-4 in Basque. Different colors used for better readability.

Spanish questions generated using the **CER framework** consistently asked students to provide a claim, support it with evidence from the text, and explain their reasoning, which involved nested questions. All eight questions included the term *afirmación* (affirmation), and key CER elements: claim, evidence, and reasoning Q5. Seven of the eight questions required students to retrieve information directly from the text. The questions were generally grammatical, clear and easy to understand. Their length varied.

Questions generated with the **divergent questioning** framework presented hypothetical scenarios and asked students to imagine and respond to new situations using single questions (Q6). They include part of the framework definition provided in the prompt with no explicit reference to the text. The language used was grammatical and natural-sounding. Question length varied, with Llama3 producing consistently short questions and GPT-4 showing more variation.

## 5.3. English

The English questions generated by the prompt based on **Bloom's Taxonomy** showed a rather stable set of tasks by cognitive level. At Analyze, 7 out of 10 questions focused on relationships between parts of a concept and asked students to analyze information (Q7). At Evaluate, half of the questions began with action verbs such as *evaluate*, *assess*, or *judge*. At Create, 5 out of 8 questions involved product-oriented tasks like designing diagrams, often framed as hypothetical or role-based prompts (Q8). Nested questions were common: in all three levels more than half of the questions were nested. The questions showed consistent specific structures per level. For example, Analyze questions from LLama3 uniformly began with *compare* and *contrast*, while GPT-4 offered more variation. All the questions were grammatical. Explicit reference to the texts was not common. Question length tended to increase with the cognitive level.

| | Question |
|---|---|
| Q3 | *¿Cómo se relacionan la fuerza centrípeta y la aceleración centrípeta en el movimiento de la Luna alrededor de la Tierra? (How are centripetal force and centripetal acceleration related in the movement of the Moon around the Earth?) (Llama3)* |
| Q4 | *Diseña un observatorio astronómico ideal para el siglo XXI, considerando factores como la ubicación, el equipamiento y la tecnología. ¿Cómo se podría aprovechar la combinación de telescopios terrestres y espaciales para avanzar en la investigación astronómica? (Design an ideal astronomical observatory for the 21st century, considering factors such as location, equipment and technology. How could the combination of ground-based and space-based telescopes be harnessed to advance astronomical research?) (Llama3)* |
| Q5 | *¿Cómo soporta la evidencia la afirmación de que El Sistema Solar se encuentra en la galaxia llamada Vía Láctea? Por favor, utiliza el razonamiento que proporciona el texto.(How does the evidence support the claim that the Solar System is located in the galaxy called the Milky Way? Please use the reasoning provided in the text.) (GPT-4)* |
| Q6 | *Cómo crees que sería el campo de la astronomía hoy en día si el telescopio no hubiera sido inventado? (What do you think the field of astronomy would be like today if the telescope had not been invented?) (GPT-4)* |

Table 5: Questions generated with Llama3 and GPT-4 in Spanish.

The English **CER**-based prompt mostly generated questions that followed the Claim-Evidence-Reasoning format. The tasks consistently ask students to justify claims with evidence and explain their reasoning through a nested question *How?* in 7 out of 8 questions (Q9). The questions generated were grammatically correct and natural. Six out of eight questions explicitly referenced the text. Question length varied.

All 15 questions generated by the prompt based on the **divergent questioning** framework were scenario-based and open-ended, requiring students to imagine themselves in specific roles such as astronauts, geologists, or ministers Q10. Many questions asked students to explore alternative outcomes, historical developments, or scientific controversies. A significant number (10 out of 15) prompted students to apply theoretical concepts to real or hypothetical situations. The questions, direct and non-nested, followed consistent patterns. Seven questions began with *Imagine you are X*, establishing role-play scenarios. *How* questions were also common. No direct reference to the source text was included. Question lengths varied between the LLMs.

| | Question |
|---|---|
| Q7 | *Analyze how seismic waves indicate the different physical states in the various layers of earth, including rigid, malleable, or fluid. (GPT-4)* |
| Q8 | *From the information provided, imagine and create a diagram of a spiral and an elliptical galaxy, including key parts discussed in the text like the nucleus and arms. Explain your diagram in terms of the text. (GPT-4)* |
| Q9 | *Identify the claim in the text about the structure and function of the Earth's layers. Then, using the evidence provided, explain your reasoning as to how this evidence supports the claim. Pay particular attention to how changes in temperature and pressure within the Earth contribute to the movements and characteristics of its layers. (GPT-4)* |
| Q10 | *Imagine you are an astronaut traveling through space and you stumble upon a galaxy that is completely different from the spiral and elliptical types we know. Describe the characteristics of this new galaxy and how it might have formed. (Llama3)* |

Table 6: Questions generated with Llama3 and GPT-4 in English.

## 6. Discussion and Conclusion

By evaluating the effectiveness of established educational frameworks such as Bloom's Taxonomy alongside CER and divergent questioning, we provide new insights into how prompts grounded on different frameworks influence high-order question generation. Our study demonstrates that state-of-the-art LLMs have the capability to generate high-order questions, although there is room for further refinement and exploration. Interestingly, their performance varies depending on the framework used, which adds diversity and complementarity to the generated questions.

A first evaluation of all generated questions by teachers revealed that a strong majority were deemed answerable. The relevance ratings revealed that most questions align well with the given text, suggesting that LLMs can generate answerable and topic-appropriate questions when guided by frameworks. Overall, just over half of the answerable questions were classified as high-order, signaling that LLMs still have a tendency to produce low-order, recall-based questions even when explicitly prompted otherwise. The results for teacher adoption are overwhelmingly positive, with most questions being considered suitable for classroom use, either directly or with modifications.

The analysis of questions generated from different educational frameworks indicates that certain approaches, particularly Bloom's Evaluate and divergent questioning, are more effective in eliciting high-order questions. Only these two prompts successfully generated create level questions, which represent the smallest portion of the generated questions. These findings align with previous research (Scaria et al., 2024), which indicates that LLMs struggle to generate create level questions, even when explicitly prompted. The comparison between Llama3 and GPT-4 reveals that both models perform similarly, with slight variations. GPT-4 shows a slight advantage in generating answerable questions, while Llama3 slightly outperforms in Relevance and Teacher Adoption for Basque. However, when considering the ability to produce high-order questions, GPT-4 appears to have a stronger capacity, particularly for English. Given that generating high-order questions is a crucial goal, GPT-4 may be considered the more effective model for this language, whereas exploring other options, even open-source ones, might be interesting for other languages, in particular, less-resourced ones.

The results across languages highlight variations in the performance of both models, suggesting that their effectiveness depends on the specific language in which they generate questions. In English, GPT-4 consistently outperforms Llama3 across most metrics, particularly in Answerability and the generation of high-order questions, with a strong presence at the analyze and evaluate levels. For Spanish, the results are more balanced, with GPT-4 performing slightly better in generating immediately answerable and adoptable questions. Both models demonstrate a strong ability to generate high-order questions, with Llama3 generating more questions at the create level. In Basque, both models show lower performance in Answerability compared to English and Spanish, but Llama3 stands out in Relevance and Teacher Adoption. Although GPT-4 generates a slightly higher number of high-order questions, Llama3 appears to be the better model for ensuring relevant and adoptable questions in this language.

A subsequent evaluation of the high-order questions confirms that the CER and divergent question-ing framework can be used alongside Bloom's Taxonomy to generate high-order questions with the help of LLMs. Even though the CER and divergent strategies helped generate only a low number of questions for Basque, in Spanish and English they provided alternative questions that would encourage students to explore different aspects of given information. Moreover, CER and divergent questioning strategies seem to generate complementary questions, as the generated questions vary both structurally and conceptually, suggesting they could be jointly used in the classroom. While both Llama3 and GPT-4 were able to generate high-order questions, notable differences in the output were observed: GPT-4 generated longer and more varied questions, be it with more grammatical issues, especially in Basque. Llama3 produced more uniform and predictable structures with fewer errors, but with less diversity when it comes to the question form, except for divergent questions.

All in all, our findings offer practical implications for both educators and researchers. We have shown that the framework in which the prompt is grounded significantly influences the type and structure of generated questions, and that integrating the CER and divergent questioning frameworks can complement Bloom's Taxonomy in supporting the development of critical thinking in the classroom. Given language-related variations and model-specific behaviors, human oversight while using AI tools remains essential. Future research should focus on refining the most relevant metrics to consider to improve evaluation consistency and to corroborate these findings on other datasets, domains and languages.

## 7. Limitations

A key challenge in our study was the low inter-annotator agreement, especially in the Answerability, Relevance and Adoption metrics. While raw agreement rates were moderate to high, Cohen's Kappa values were low, likely due to a skewed (positive) response distribution. To examine this, we used Gwet's AC1/AC2, which revealed a slight Kappa paradox in the Answerability and Adoption metrics. There are multiple reasons why agreements might be low, from the subjectivity of the relevant metrics to the limited set of question items and evaluators involved in the experiments. Given the promising results obtained, it would be interesting to enlarge the scope of the study in future work. Also, the use of Bloom's Taxonomy for the classification of the cognitive level of the questions might somehow favor the results for this particular framework, even when it can be argued to be a generalizable scale that teachers know well, as reported by our participants.

Our findings are model-specific as we only relied on GPT-4 and Llama3 for question generation. Investigating additional models could provide a broader understanding of LLMs' capabilities in educational settings. This seems to be particularly relevant for different languages, as models show different behavior probably because of the representation of data during training. Also, the source texts were limited to the domain of physics. This may limit the generalizability of the findings to other subject areas and would be worth expanding.

## 8. Acknowledgements

We would like to thank the anonymous reviewers for their valuable insights. This research was supported by the University of the Basque Country (PIF20/154 UPV/EHU 2020); the LUMINOUS project (HE-CL4-DIGITAL23/02 (101135724)); the MOLVI project (PID2024-157855OB-C32), funded by MICIU/AEI/10.13039/501100011033 and co-funded by FEDER, EU; the project "Desarrollo de Modelos ALIA" (Resolución SE-DIA 19.08.2024), within the framework of the National Language Technologies Plan (ENIA 2024), funded by MTDFP, PRTR, and the European Union-NextGenerationEU; the CRITICS project (PCI2025-167239-2), funded by MICIU/AEI/10.13039/501100011033 and co-funded by the European Union; the Grant DeepThought (PID2024-159202OB-C21) funded by MICIU/AEI /10.13039/501100011033 and by ERDF, EU and the DAWN project funded by the Special Research Fund of Ghent University under grant number BOF.STG.2022.0012.01.

## 9. Bibliographical References

Donna Rae Clasen and Robert E. clasen. 1990. Teachers trackle thinking.

Zulmaris Diaz, Michael Whitacre, J Joy Esquierdo, and Jose A Ruiz-Escalante. 2013. Why did i ask that question? bilingual/esl pre-service teachers' insights. *International Journal of instruction*, 6(2).

Sabina Elkins, Ekaterina Kochmar, Jackie CK Cheung, and Iulian Serban. 2024. How teachers can use large language models and bloom's taxonomy to create educational quizzes. In *Proceedings of the AAAI Conference on Artificial Intelligence*, volume 38, pages 23084–23091.

Sabina Elkins, Ekaterina Kochmar, Iulian Serban, and Jackie CK Cheung. 2023. How useful are educational questions generated by large language models? In *International Conference on Artificial Intelligence in Education*, pages 536–542. Springer.

Alvan R Feinstein and Domenic V Cicchetti. 1990. High agreement but low kappa: I. the problems of two paradoxes. *Journal of clinical epidemiology*, 43(6):543–549.

Meredith D Gall. 1970. The use of questions in teaching. *Review of educational research*, 40(5):707–721.

James J Gallagher and Mary Jane Aschner. 1963. A preliminary report on analyses of classroom interaction. *Merrill-Palmer Quarterly of Behavior and Development*, 9(3):183–194.

Kilem Li Gwet. 2008. Computing inter-rater reliability and its variance in the presence of high agreement. *British Journal of Mathematical and Statistical Psychology*, 61(1):29–48.

DR Krathwohl. 2002. A revision bloom's taxonomy: An overview. *Theory into Practice*.

Joanne Gainen Kurfiss. 1988. Critical thinking: Theory, research, practice, and possibilities. asheeric higher education report no. 2, 1988.

David R McNamara. 1981. Teaching skill: the question of questi. *Educational Research*, 23(2):104–109.

Katherine L McNeill, David J Lizotte, Joseph Krajcik, and Ronald W Marx. 2006. Supporting students' construction of scientific explanations by fading scaffolds in instructional materials. *The journal of the Learning Sciences*, 15(2):153–191.

Robert T Pate and Neville H Bremer. 1967. Guiding learning through skilful questioning. *The Elementary School Journal*, 67(8):417–422.

Luise B Savage. 1998. Eliciting critical thinking skills through questioning. *The Clearing House*, 71(5):291–293.

Nicy Scaria, Suma Dharani Chenna, and Deepak Subramani. 2024. Automated educational question generation at different bloom's skill levels using large language models: Strategies and evaluation. In *International Conference on Artificial Intelligence in Education*, pages 165–179. Springer.

Suna-Şeyma Uçar, Itziar Aldabe, Nora Aranberri, and Ana Arruarte. 2024. Exploring automatic readability assessment for science documents within a multilingual educational context. *International Journal of Artificial Intelligence in Education*, pages 1–43.